%% file: main.tex
\documentclass[letterpaper, 10 pt, conference]{ieeeconf}  % Comment this line out if you need a4paper

\IEEEoverridecommandlockouts                              % This command is only needed if 
\usepackage{graphics}
\usepackage{multirow}
\usepackage{graphics} % for pdf, bitmapped graphics files
\usepackage{epsfig} % for postscript graphics files
\usepackage{times} % assumes new font selection scheme installed
\usepackage{amsmath} % assumes amsmath package installed
\usepackage{amssymb}  % assumes amsmath package installed

\usepackage{array}
\usepackage{booktabs}

\usepackage[hidelinks]{hyperref}
\usepackage{cite}

\usepackage{xcolor,pifont}
\usepackage{color}
\usepackage[nolist]{acronym}
\usepackage{amsmath}
\usepackage{multirow}
\usepackage{microtype}

\newcommand{\Figref}[1]{Figure~\ref{#1}}  % beginning of sentence
\newcommand{\figref}[1]{Fig.~\ref{#1}}    % somewhere

\newcommand{\tabref}[1]{Table~\ref{#1}}

\newcommand{\eqnref}[1]{Eq.~\ref{#1}} % Eq. (1)
\title{\LARGE \bf
DFM: Deep Fourier Mimic for Expressive Dance Motion Learning
}

\title{\LARGE \bf
Learning Quiet Walking for a Small Home Robot
}

\author{Ryo Watanabe$^{1}$$^{,}$$^{2}$, Takahiro Miki$^{1}$, Fan Shi$^{1}$$^{,}$$^{3}$, Yuki Kadokawa$^{1}$$^{,}$$^{4}$, Filip Bjelonic$^{1}$, \\
Kento Kawaharazuka$^{1}$$^{,}$$^{5}$, Andrei Cramariuc$^{1}$ and Marco Hutter$^{1}$% <-this % stops a space
\thanks{$^{1}$Ryo Watanabe, Takahiro Miki, Fan Shi, Yuki Kadokawa, Filip Bjelonic, Kento Kawaharazuka, Andrei Cramariuc, and Marco Hutter are with the Robotic Systems Lab, Department of Mechanical Engineering, ETH Zurich, Switzerland.}
\thanks{$^{2}$Ryo Watanabe is at Sony Group Corporation, Japan}%
\thanks{$^{3}$Fan Shi is at National University of Singapore}%
\thanks{$^{4}$Yuki Kadokawa is at Nara Institute of Science and Technology, Japan}%
\thanks{$^{5}$Kento Kawaharazuka is at The University of Tokyo, Japan}%
}

\begin{document}

\maketitle
\thispagestyle{empty}
\pagestyle{empty}

%%%%%%%%%%%%%%%%%%%%%%%%%%%%%%%%%%%%%%%%%%%%%%%%%%%%%%%%%%%%%%%%%%%%%%%%%%%%%%%%
\begin{abstract}
As home robotics gains traction, robots are increasingly integrated into households, offering companionship and assistance. Quadruped robots, particularly those resembling dogs, have emerged as popular alternatives for traditional pets. However, user feedback highlights concerns about the noise these robots generate during walking at home, particularly the loud footstep sound.
To address this issue, we propose a sim-to-real
based \ac{rl} approach to minimize the foot contact velocity highly related to the footstep sound. Our framework incorporates three key elements: learning varying PD gains to actively dampen and stiffen each joint, utilizing foot contact sensors, and employing curriculum learning to gradually enforce penalties on foot contact velocity. Experiments demonstrate that our learned policy achieves superior quietness compared to a \ac{rl} baseline and the carefully handcrafted Sony commercial controllers. Furthermore, the trade-off between robustness and quietness is shown. This research contributes to developing quieter and more user-friendly robotic companions in home environments.
% This paper addresses this issue by introducing a reinforcement learning (RL) based locomotion policy learned in a simulator and successfully transferred to the real home robot, aibo. Our approach includes three key elements: learning varying PD gains, using foot contact sensors, and utilizing curriculum learning. Evaluation of the resulting locomotion controller shows that it achieves quieter locomotion than Sony's commercial locomotion and our RL baseline locomotion, as measured by sound magnitude within the audible range. This research opens avenues for quieter and thus more user-friendly robotic companions in home environments.
%Our contributions include the first implementation of a quiet walking policy in RL for quadrupedal robots, deployment on a real consumer-grade small platform, and superior performance compared to baseline RL and Sony commercial controllers. => repetitive. 
\end{abstract}

\input{sections/1-intro}

\input{sections/2-related}
\input{sections/3-method}

\input{sections/4-result}

\input{sections/5-discussion}
\input{sections/6-conclusion}

\input{acronyms}

%%%%%%%%%%%%%%%%%%%%%%%%%%%%%%%%%%%%%%%%%%%%%%%%%%%%%%%%%%%%%%%%%%%%%%%%%%%%%%%%
\iffalse
\section*{APPENDIX}

\begin{itemize}
    \item Furter resources on mono, stereo and event camera latency.
    \item Datasheet extracts to back our claims, if needed.
    \item Elaboration of the testing setup.
\end{itemize}
\fi

\section*{ACKNOWLEDGMENT}
The authors would like to thank Hiroyuki Izumi, Kensuke Kitamura, Ichitaro Kohara, Fumitaka Joo, Toshihisa Sambommatsu, Takuma Morita, and Yuichiro To at Sony Group Corporation for software integration help. 
Thanks to Yuntao Ma, Mayank Mittal and Jonas Frey at ETH Zurich for discussion about reinforcement learning.  

\clearpage

%%%%%%%%%%%%%%%%%%%%%%%%%%%%%%%%%%%%%%%%%%%%%%%%%%%%%%%%%%%%%%%%%%%%%%%%%%%%%%%%

\bibliographystyle{IEEEtran}
\bibliography{main} 

\end{document}

%% file: sections/1-intro.tex
\section{INTRODUCTION}
% Introduction into the topic
Home robotics is a rapidly growing trend driven by advancements in actuators, sensors, and artificial intelligence. These developments have allowed home robots to perform a wider range of tasks and interact with humans in more meaningful ways~\cite{aibo, lovot, qoobo, disney_learning}. 
For example, aibo~\cite{aibo}, a robotic pet introduced by Sony, is capable of learning and developing its own unique personality through interactions with its home environment and owner. This emotional connection with robotic pets demonstrates the potential for robots to play a significant role in our daily lives~\cite{companionAIBO2003, tanaka2021pilot, AIBOasDog2003, KANG2020207}.

\begin{figure}[!t]
    \centering
    % \vspace{5pt}
    \includegraphics[width=\linewidth]{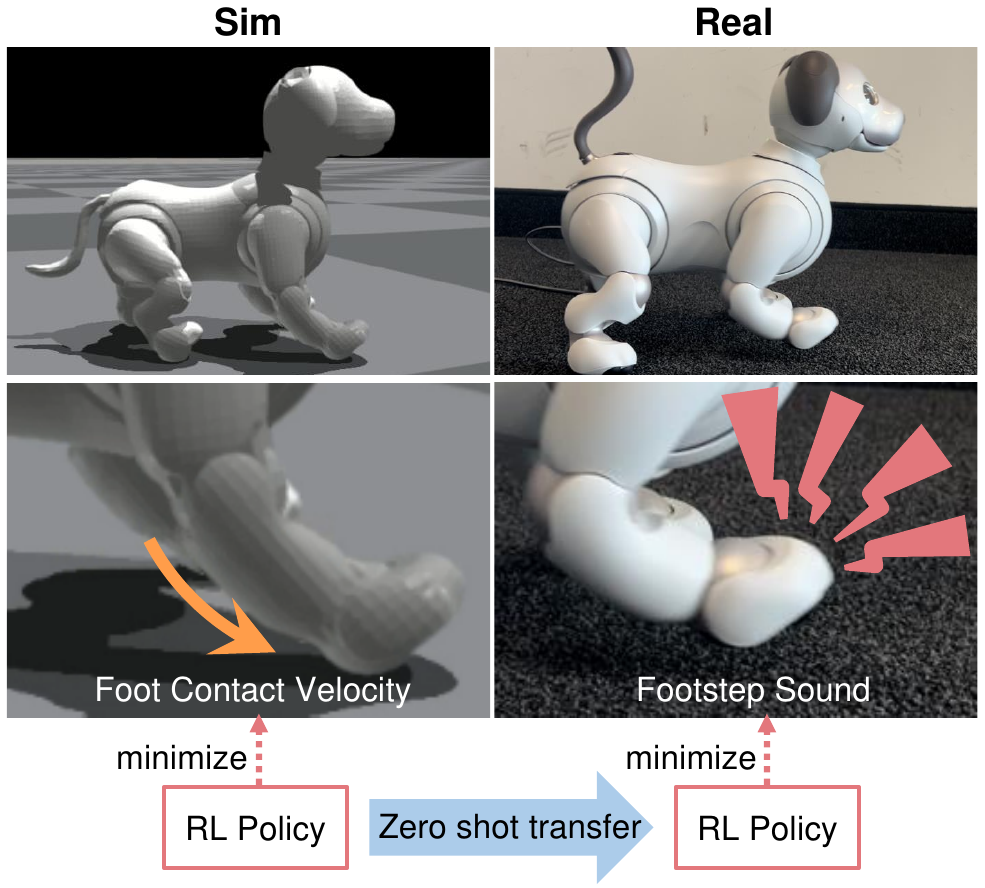}
    \caption{For the aibo small home robot pictured above, we design a sim-to-real based \ac{rl} approach to minimize the foot contact velocity in the physics simulator, which highly correlates with the footstep sound in the real world to achieve quiet walking. Project webpage: \url{https://sony.github.io/QuietWalk/}}
    \label{fig:quiet_walking_concept}
    \vspace{-2ex}
\end{figure}

\begin{figure*}
    \centering
    \vspace{10pt}
    \includegraphics[width=\linewidth]{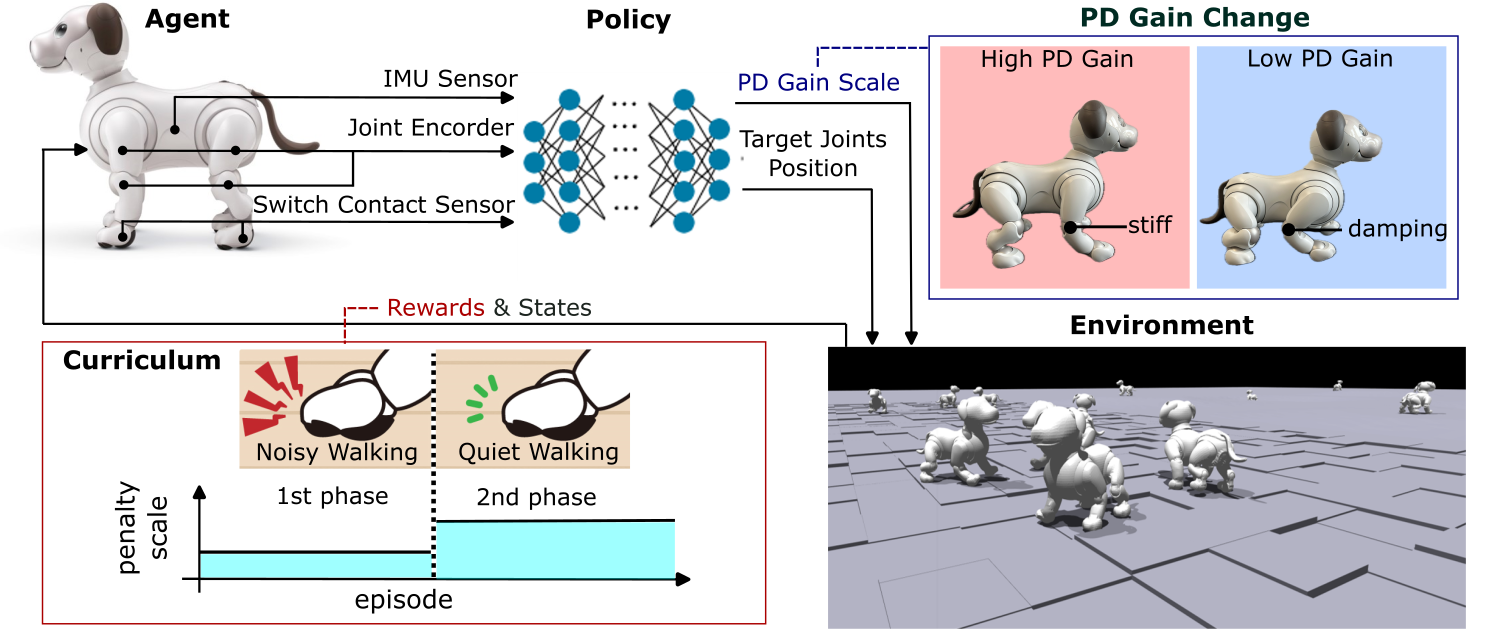}
    \caption{System overview of the \ac{rl} training framework for learning quiet locomotion with aibo. The agent, aibo, uses its IMU, joint encoders, and switch contact sensors to compute observations. As an action for 12 joints, the policy outputs target joint position and the joint PD gain scale, which enables the tracking of the target joints with high PD gain and the dampening of the joints with low PD gain. On the right is the Isaac Gym~\cite{Makoviychuk2021-th} simulation environment, which we leverage for parallel training on GPUs of multiple agents. The reward scales are divided into two phases, where aibo first learns just to walk, and then in the second phase, it adapts its walk to be quieter.}
    \label{fig:system_overview}
    \vspace{-2ex}
\end{figure*}

Reinforcement learning (RL) has recently been applied to robotic locomotion, significantly improving robustness in a variety of terrains~\cite{anymal_terrain, anymal_perceptive, Choi2023-cf, Wu2023-nz, anymal_dtc, legged_gym}. While these advances have proven beneficial for outdoor use, bringing similar robustness and adaptability to home robots remains an important goal. However, much of the research in \ac{rl} based legged locomotion has prioritized robustness and efficiency, with little focus on reducing the footstep noise generated during walking.
In home environments, where quiet operation is critical, reducing walking noise is essential. For example, one of the main concerns of aibo users is that the walking sound is too loud. This feedback highlights the need for quieter locomotion policies in home robots.

In this work, we focus on minimizing footstep noise by using sim-to-real based \ac{rl} to reduce foot contact velocity, which serves as a proxy for footstep sound reduction, as shown in \figref{fig:quiet_walking_concept}. To achieve this, we introduce three key elements aimed at effectively reducing foot contact velocity during locomotion. First, we allow the \ac{rl} network to dynamically adjust the PD gains of the actuators, allowing for better joint damping and stiffening. Second, we incorporate foot switch contact sensors that detect when the robot's foot touches the ground to selectively dampen joints during the contact phase and stiffen them during the stance phase for body support. Third, we employ a curriculum learning approach that initially trains the robot on basic locomotion and progressively applies penalties to noisy walking such as foot contact velocity to ensure quieter walking.

We validate our approach through hardware experiments with Sony's quadruped robot aibo. In these experiments, we show that foot contact velocity correlates with footstep noise, and our method produces the quietest locomotion compared to both Sony's carefully hand-tuned controllers and other \ac{rl} baselines. In addition, we explore the trade-off between robustness and quietness in locomotion and show that this balance can be adjusted using \ac{dr} parameters.

In summary, our main contributions are as follows: 
\begin{itemize} 
\item Demonstrate superior quiet locomotion in a small home robot by minimizing physical parameters, such as foot contact velocity, that correlate with footstep sound using a sim-to-real \ac{rl} approach. 
\item Develop an RL framework that integrates three core elements: adaptive PD gains, foot contact sensors, and curriculum learning, resulting in an effective policy for quiet walking. 
\item Identify and demonstrate the trade-off between robustness and quietness in robot locomotion, with \ac{dr} parameters providing a way to balance these competing objectives. 
\end{itemize}

%% file: sections/2-related.tex
\section{RELATED WORK}
\subsection{Reinforcement Learning Locomotion}
RL has been successfully used to generate robust locomotion policies for quadrupedal robots~\cite{anymal_terrain, anymal_perceptive, Choi2023-cf, Wu2023-nz}. 
All of the previously mentioned works focus on robustness but ignore the user-interaction aspect of the locomotion policy, namely the amount of noise the robot generates when moving around. 
% The resulting policies are often loud and stompy, mainly because they do not prioritize quietness. 
Even locomotion policies designed to be energy efficient do not inadvertently reduce the amount of noise they produce, as seen in the work of Yang \textit{et al.}~\cite{energyefficient_learning_locomotion}. 
Other works on energy efficient gaits~\cite{energyefficient_amp_locomotion} or imitation learning-based gaits do not address or evaluate the impact noise of the learned policy~\cite{fuchioka2023opt, reske2021imitation}. 
While many of these works show promising results for robustness and efficiency, they are not designed to suppress loudness, which we aim to solve.

\subsection{Sound Engineering in Robotics} Sound plays a crucial role in how humans perceive and interact with various products.
Extensive engineering efforts have been made in the automotive industry~\cite{parizet2008analysis} to optimize the acoustic experience for drivers. Recently, the robotics industry has also recognized the importance of sound engineering, with research focusing on robotic arms~\cite{tennent2017good, trovato2018sound}, bipedal robots~\cite{stelthwalking}, rolling robots~\cite{quietwalkingrobot}, and servomotors~\cite{servo_sound}. These studies have investigated how the generated sounds influence human comfort and perception during human-robot interactions, leading to the development of more pleasant and even cute motion patterns~\cite{arm_motion_sound_study}. 
However, previous works have not thoroughly explored the acoustic impact of quadruped robot footsteps. 
This paper aims to address this gap by focusing on evaluating and reducing the sound generated by the footsteps of quadruped robots. 

%% file: sections/3-method.tex
\section{METHOD}
To minimize foot contact velocity in the physics simulator correlated to the footstep sound in the real, we propose a sim-to-real deep \ac{rl} framework which synergistically incorporates three key components as shown in \figref{fig:system_overview}.

The training network architecture consists of three fully connected layers, each with 128 hidden units and \ac{elu} activation functions, and utilizes the \ac{ppo}~\cite{ppo}.
Both the actor and critic networks are implemented as \ac{mlp}, adhering to the architecture and PPO hyperparameters established in prior work~\cite{legged_gym}.
The simulation and control loop frequencies are set to 400 Hz and 100 Hz, respectively, using the Isaac Gym framework~\cite{Makoviychuk2021-th}.

\begin{table}
\caption{Policy Observation}
\label{table:observation}
\begin{center}
\begin{tabular}{ccccc}
\toprule
\textbf{Entry} & \textbf{Symbol} & \textbf{Noise} & \textbf{Units} & \textbf{Dimensions} \\
 & &  \textbf{level}\\
\midrule
Joint positions & $\boldsymbol{\varphi}$ & 0.01 & $\rm{rad}$ & $12$ \\
Joint velocities & $\boldsymbol{\dot{\varphi}}$ & 1.5 & $\rm{rad/s}$ & $12$ \\
Last action (joint position) & $\boldsymbol{a}^*$ & 0.0 & $\rm{rad}$ & $12$\\
Last action (PD gain scale) & $\boldsymbol{gain}^*$ & 0.0 & $\rm{-}$ & $12$\\
Foot contact state & $\boldsymbol{f_c}$ & 0.0 & $-$ & $4$\\
Gravity orientation & $\boldsymbol{g}$ & 0.05 & $m/s^2$ & $3$\\
\bottomrule
\end{tabular}
\end{center}
\end{table}

\subsection{Actions}
As shown in \figref{fig:system_overview}, the policy outputs two types of actions for the 12 leg joints: target joint positions and PD gain scales. $\boldsymbol{a}^*$ and $\boldsymbol{gain}^*$ represent target joint positions and PD gain scales, respectively. 
We chose target joint position and PD gain scale as actions instead of directly estimating torque for two primary reasons. First, previous work~\cite{peng2017learning} reported that torque control as an action wasn't efficient for learning locomotion. Secondly, previous work~\cite{bogdanovic2020learning} found that estimating target joint positions and PD gains is more suitable for sensitive contact tasks that require soft ground contact.

In our approach, the proportional $P_i$ and derivative $D_i$ gains are calculated using the following formulas:
\begin{equation}
\begin{split}
P_i &= P^* + \alpha\sigma({x}_{i}) \\
D_i &= D^* + \beta\sigma({x}_{i})
\end{split}
\label{eq:pd_gain_scale}
\end{equation}
where $\sigma$ is the sigmoid function, $\alpha$ and $\beta$ are scaling factors, $P^*$ and $D^*$ are the nominal gains, and $i$ represents each leg joint. The specific parameter values are $\alpha = 4.0$, $\beta = 0.02$, $P^* = 3.0$, and $D^* = 0.03$.
Our locomotion policy outputs only one gain scale factor $x_i$ for both the proportional $P_i$ and derivative $D_i$ parts of the PD controller. 
While this approach of outputting a single gain scale from the policy is consistent with the previously mentioned work~\cite{bogdanovic2020learning}, this method is the first application to locomotion tasks in quadruped robots to explore the quietness.

\subsection{Observations}
Policy observations and noise levels are detailed in Table~\ref{table:observation}. One of the key elements of our approach is utilizing the switch contact sensor to determine when to stiffen the joint from its soft state during the contact phase. Since aibo is a consumer-grade small robot designed to be affordable for the average household, it does not have force and torque sensors. Without these sensors, there is a traditional way to estimate external force to detect contact using model-based methods \cite{traditional_torque_control}. However, such methods are not sufficiently accurate to determine when a small robot contacts the ground, as it is difficult to precisely identify friction and develop an accurate robot model for commercially available mass production robots. Instead of estimating external force, we utilized switch contact sensors to detect the contact directly. Although the switch contact sensors provide binary output, they are still useful for the contact sensitive task such as quiet walking.

\begin{table*}
\caption{Reward functions and scales at each curriculum learning phase}
\label{table:reward}
\begin{center}
\begin{tabular}{lllcc}
\toprule
\textbf{Category} & \textbf{Reward} & \textbf{Definition} & \textbf{Scale in noisy} & \textbf{Scale in quiet} \\
 & & & \textbf{walking phase} & \textbf{walking phase} \\
\midrule
\multirow{2}{*}{\textbf{Task Rewards}} & Linear velocity tracking & $\exp(-\frac{1}{0.06}\|\mathbf{v}_{b,xy}^* - \mathbf{v}_{b,xy}\|^2 )$ & 1.0 & 1.0  \\
 %& Angular velocity tracking & $\exp(-\frac{1}{0.06}\|\mathbf{w}_{b,z}^* - \mathbf{w}_{b,z}\|^2)$ & 1.0 & 1.0 \\
 & Angular velocity tracking & $\exp(-\frac{1}{0.06}\|{w}_{b,z}^* - {w}_{b,z}\|^2)$ & 1.0 & 1.0 \\
\midrule
\multirow{8}{*}{\textbf{Penalty Rewards}} & Joint torque & $\|\boldsymbol{\tau}\|^2$ & -0.015 & -0.015  \\
 & Base linear velocity z & $\|v_{b,z}\|^2$ & -3.0 & -3.0 \\
 & Base orientation & $\|\boldsymbol{\vartheta}_{b,xy}\|^2$ & -5.0 & -5.0 \\
 & Base angular velocity & $\|\boldsymbol{\dot\vartheta}_{b,xy}\|^2$ & -0.05 & -0.05 \\
 & Foot slippage & $\|\boldsymbol{v}_{f,xy}\|^2$ & -0.15 & -0.15 \\
 & Self-collisions & $n_{collision}$ & -10.0 & -10.0 \\
 & Foot air time & $\sum_i (\boldsymbol{t}_{f,air} - 0.2)$ & 2.0 & 2.0 \\
 & Joint target difference & $\|\boldsymbol{a}^*_{t-1} - \boldsymbol{a}^*_{t}\|^2$ & -0.02 & -0.02 \\
 & PD gain scale difference & $\|\boldsymbol{gain}^*_{t-1} - \boldsymbol{gain}^*_{t}\|^2$ & -0.005 & -0.005 \\
\midrule
\multirow{3}{*}{\textbf{Noisy Walking Penalties}}  & Foot contact velocity & $\|\boldsymbol{v}_{f,xyz}\|^2$ & -5.0 & -25.0 \\
& Joint acceleration & $\|\boldsymbol{\ddot\varphi}\|^2$ & -2e-7 & -4e-7 \\
 & Base angular acceleration & $\|\boldsymbol{\ddot\vartheta}_{b,xy}\|^2$ & -5e-5 & -1e-4 \\
\bottomrule
\end{tabular}
\end{center}
\end{table*}

\subsection{Rewards}
Reward functions and their relative scales are described in Table~\ref{table:reward}. 
For task rewards, $\boldsymbol{v}_{b}$, $\boldsymbol{v}_{b}^*$, $\boldsymbol{w}_{b}$, and $\boldsymbol{w}_{b}^*$ denote the measured and desired linear velocities in the base frame and the measured and desired angular velocities in the base frame, respectively.
Penalty rewards help suppress inappropriate or jerky motions. $\boldsymbol{\varphi}$, $\boldsymbol{\tau}$, $\boldsymbol{v}_{f}$, ${n}_{collision}$ and $\boldsymbol{t}_{f,air}$ represent joint positions, joint torques, foot velocities, number of collisions and airtime of foot, respectively.  

We incorporate three additional terms related to the footstep sound for quiet walking. 
Biomechanics research indicates that footstep sound correlates with foot contact velocity~\cite{acoustic_footstep}. When the foot contacts the ground, the kinetic energy, $\frac{1}{2} m \mathbf{v}_{f,xyz}^2$, is converted into various forms of energy, including sound energy which produces the footstep sound. 
While foot contact velocity is the dominant parameter related to footstep sound, we also include joint acceleration and base angular acceleration penalties to support this objective. We observe that sudden changes in joint angles and base orientation often follow loud step sounds, as noisy walking causes the robot to bump into the ground.

\subsection{Curriculum Learning}
Initial attempts at training without curriculum learning proved unsuccessful, resulting in either non-convergent training or the aibo learning to remain stationary, a strong local minimum that minimizes noise but fails to achieve the desired locomotion.
To address these challenges, we implemented a curriculum learning approach~\cite{curriculum1}. Inspired by a recent study proposing the division of a training episode into multiple stages to tackle the explore vs exploit dilemma~\cite{tuyls2022multi, hartmann2024deep}, we divide the training episode into two distinct stages: noisy walking and quiet walking stage. 

During the noisy walking phase, we applied lower weights to the noise-related penalties, as shown in Table~\ref{table:reward}. This strategy allowed the robot to prioritize learning basic locomotion and velocity command tracking, even if significant footstep sounds are generated. 
The transition to the quiet walking phase is triggered when the agents achieve the sum of episodic reward about linear and angular velocity tracking is more than $1.5$.
In the quiet walking phase, we increased the penalties associated with foot contact velocity, base angular acceleration, and joint acceleration.  
Given that foot contact velocity is the most critical parameter for footstep sound reduction, its penalty was increased 5 times compared to the noisy walking phase, while other penalties were only doubled as shown in Table~\ref{table:reward}. 
By combining reward shaping with this two-phase curriculum learning approach, we successfully enabled the robot to develop a quiet locomotion policy that minimizes the footstep sound while preserving its capacity to execute velocity commands effectively.

\begin{table}[!t]
\caption{Randomized Simulation Parameters}
\label{table:domainrandomizationparameters}
\begin{center}
\begin{tabular}{cccc}
\toprule
\textbf{Randomized Variables} & \textbf{unit} & \textbf{min} & \textbf{max} \\
\midrule
Robot base mass & kg & -0.4 & 0.4 \\
Velocity disturbance & m/s & -0.2 & 0.2 \\
External force & N & 0.0 & 0.4 \\
External torque & Nm & 0.0 & 0.1 \\
Terrain height & m & 0.002 & 0.01 \\
Friction coefficient & [-] & 0.4 & 0.7 \\
\bottomrule
\end{tabular}
\end{center}
\end{table}

\subsection{Domain Randomization}
We employ \ac{dr}~\cite{domainrandomization1} to enhance the sim-to-real transfer of the policy. This involves procedurally generating various parameters in the simulation environment, such as the height of the steps, the ground friction coefficient, and the robot base mass. Additionally, we utilize disturbances during the episodes by applying an impulse signal to the robot’s base velocity every 4 seconds and random external forces and torques to the base. The ranges for the randomized parameters are shown in \tabref{table:domainrandomizationparameters}.

The extent of randomization significantly influences the resulting policy's characteristics. While more extensive \ac{dr} typically leads to a more robust gait, it can potentially compromise the quietness of locomotion. Previous research on \ac{rl}-based racing drones~\cite{song2023reaching} has mentioned the trade-off between robustness and performance depending on the degree of \ac{dr}. Similarly, for quiet walking, it is crucial to strike an optimal balance when determining the level of \ac{dr}.

% We employ \ac{dr}~\cite{domainrandomization1} to enhance the sim-to-real transfer of the policy, as detailed in Table~\ref{table:domainrandomizationparameters}.
% The extent of randomization significantly influences the resulting policy's characteristics. While more extensive \ac{dr} typically leads to a more robust gait, it can potentially compromise the quietness of locomotion. Previous research on \ac{rl}-based racing drones~\cite{song2023reaching} has mentioned the trade-off between robustness and performance depending on the degree of \ac{dr}. Similarly, for quiet walking, it is crucial to strike an optimal balance when determining the level of \ac{dr}.

%% file: sections/4-result.tex
\section{Experiments and Results}
We evaluate the quietness by using the microphone in the robot depending on the measured base velocity. By showing that the selective reward such as the foot contact velocity, base angular acceleration and joint angular acceleration at the proposed policy are smaller than the baseline policy, we prove those observable parameters in a physics simulator correlate to the footstep sound in the real. As part of our ablation study, we demonstrate the trade-off between robustness and quietness by validating three key factors of our framework and the effectiveness of \ac{dr}.  
\begin{figure}[!t]
    \centering
    \vspace{-20pt}
    \includegraphics[trim={0 0 0 0}, width=\linewidth]
    {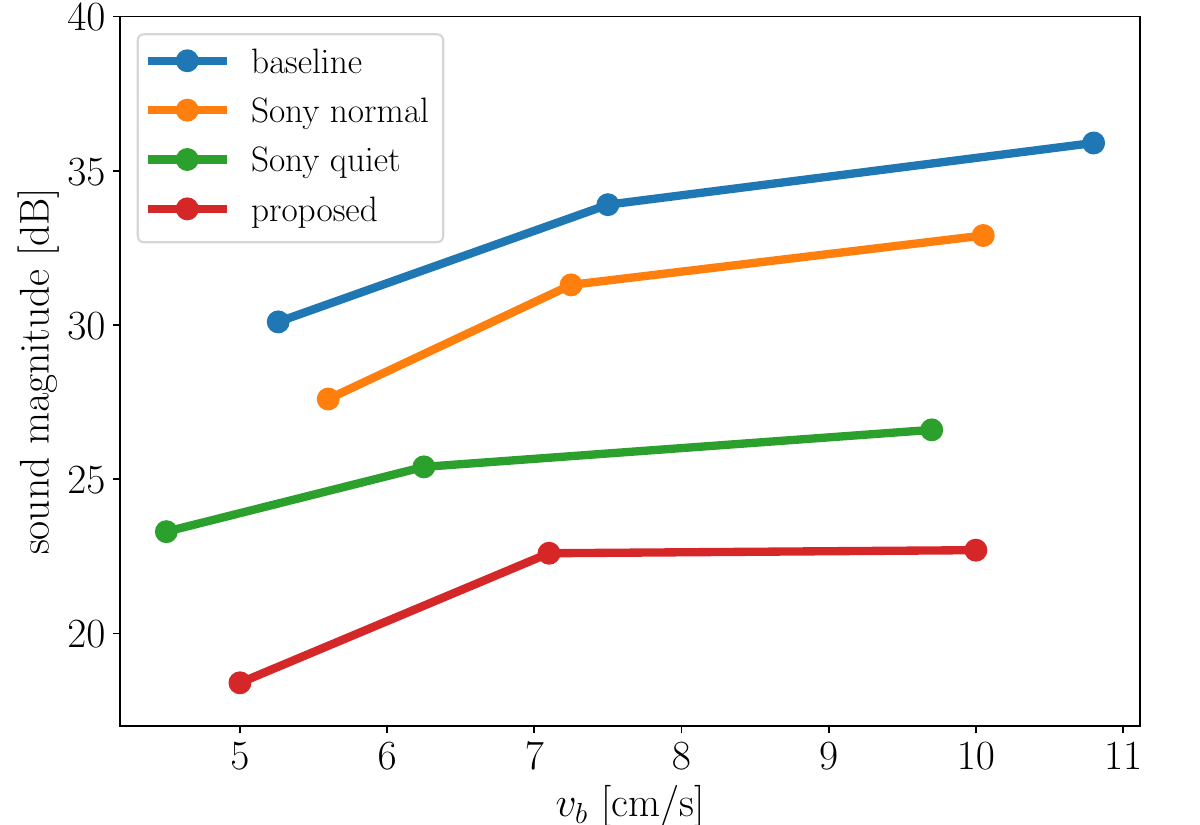}
    \caption{Comparison of the average sound magnitude between $20$ Hz and $20$ kHz for the measured base velocity of aibo. The baseline and propose are RL policies without and with our three key factors for each. Sony normal and Sony quiet are the commercial controllers provided by Sony.}
    \label{fig:sound_eval_vel}
    \vspace{-2ex}
\end{figure}

\subsection{Quietness Evaluation}
\label{subsec:quietness_eval}
We evaluate the performance of our proposed quiet \ac{rl} walking policy and compare it to three other methods: a baseline \ac{rl} policy and two commercial locomotion controllers provided by Sony. 
The baseline \ac{rl} policy is trained using the standard rewards from the previous work~\cite{legged_gym} which we tuned to adapt to the small-scale robot, aibo. Unlike our proposed policy, the baseline \ac{rl} policy does not output PD gain scale, does not include rewards for noisy walking penalties, and does not employ curriculum learning during training. 
Sony's two baseline controllers are designed for normal walking and quiet walking, respectively. 
As these are proprietary algorithms, we compare our method to them without knowledge of their internal workings.

To evaluate the sound level of each locomotion policy, we use one of the microphones on the rear side of aibo's head, which has a sampling frequency of $48$ kHz. We focus our analysis on the standard human audible range between $20$ Hz and $20$ kHz~\cite{neuroscience}, as this range is most relevant for assessing perceived noise levels. In this experiment, all locomotion controllers are set to track the same base velocity moving forward for the comparison. We preprocess the data using a Hamming window and then use Welch's method~\cite{welch1967use} to perform the frequency analysis with \acp{fft}. We use a window size of $4096$ samples corresponding to approximately $85$ milliseconds. For the power analysis, we use as a reference a $10^{-10}$ amplitude sinusoid for each frequency bin.

The sound level of walking increases with higher commanded base velocities, as the robot has less time to slow its feet before contact while maintaining the commanded velocity. Therefore, we experimented to measure the average sound level for each policy at different measured linear velocity commands. It is important to note that the microphone is approximately $10$ cm away from the feet, whereas the sound level would be much lower for a human standing further away. We only use the human audible frequency range and a $1$ kHz sinusoidal reference signal with an amplitude of $1.0$. \Figref{fig:sound_eval_vel} shows that the quietness of proposed method outperforms the other methods such as the baselines and Sony commercial controllers at different measured velocities. 
% We achieve an average of $10.9$ dB noise reduction with respect to the baseline \ac{rl} policy. We also achieve a $1.73$ dB reduction in noise compared to the best baseline, i.e. Sony's quiet walk, which represents a substantial improvement in quietness, equivalent to a $49$\% reduction.

\begin{table}[!t]
\caption{Analysis of Noisy Walking Penalties at Simulator}
%  for \\ Validating the Correlation of Footstep Sound at Real
\label{table:reward_analysis}
\begin{center}
\begin{tabular}{cccc}
\toprule
\textbf{Physical Term} & \textbf{Unit} & \textbf{Baseline} & \textbf{Proposed} \\
\midrule
Foot contact velocity & $\rm{m/s}$ & 0.417 & 0.123 \\
Joint acceleration & $\rm{rad/s^2}$ & 114.3 & 76.7 \\
Base angular acceleration & $\rm{rad/s^2}$ & 57.2 & 23.7\\
\bottomrule
\end{tabular}
\end{center}
\end{table}

\subsection{Analysis of Noisy Walking Penalties Rewards}
To validate our approach of minimizing noisy walking penalties in the simulator for reducing aibo's footstep sound in the real world, as illustrated in \figref{fig:quiet_walking_concept}, we compared the noisy walking penalties between the baseline and proposed policies. 
~\tabref{table:reward_analysis} presents the average values for 10 seconds calculated using the equations in Table~\ref{table:reward} in the Isaac Gym simulator.
The policies used for the baseline and proposed are those employed in \figref{fig:sound_eval_vel}. 
As shown in \tabref{table:reward_analysis}, our analysis reveals that the proposed policy, which produces quieter walking compared to the baseline, exhibits lower values across all observed parameters. As mentioned in subsection \ref{subsec:quietness_eval}, the results indicate that the footstep sounds generated by the proposed policy are quieter than those of the baseline.
This finding suggests a correlation between the real-world footstep sound and the noisy walking penalties in the simulator. Consequently, these penalties such as foot contact velocity, joint acceleration, and base angular acceleration effectively enable aibo to walk quietly in the real world through our sim-to-real transfer approach.

\begin{figure}
    \centering
    % \vspace{-5pt}
    \includegraphics[trim={0 0 0 0}, width=\linewidth]{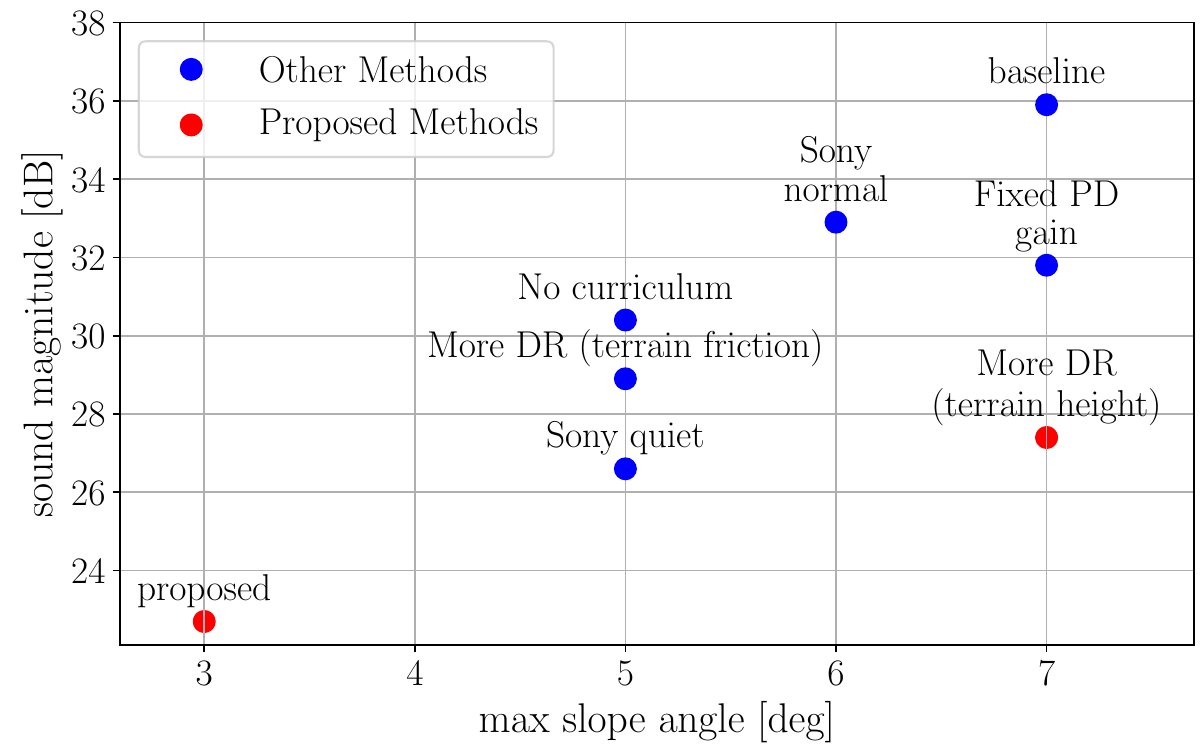}
    \caption{Trade-off between robustness and quietness are shown. The experiment evaluates robustness by having aibo climb a slope. Sound magnitude is calculated using the same method as in \figref{fig:sound_eval_vel}. The controllers such as \ac{rl} baseline, ablation test, SONY commercial controllers, and \ac{rl} proposed are shown.}
    \label{fig:tradeoff}
    \vspace{-2ex}
\end{figure}

\subsection{Trade-off Between Robustness and Quietness}
To assess robustness, we designed an experiment where aibo had to traverse a 0.5 m long slope within 20 seconds. We varied the slope angle and used this value as a metric to compare robustness, assuming that adaptability to unknown environments is one of the important aspect of robustness. We measured sound magnitude within the audible range using the method described in subsection \ref{subsec:quietness_eval}. We prepared policies including our proposed method, baseline, Sony commercial controllers, and ablation study conditions for the framework, which consists of three key factors.

\Figref{fig:tradeoff} shows that the loudest baseline policy can overcome the steepest slope of 7 degrees. Although our proposed method is the quietest among all settings, its robustness is the lowest. This observation suggests a trade-off between quietness and robustness, as quieter policies struggle to overcome steeper slopes.

We conducted an ablation study by training aibo in eight different settings. First, we evaluated policies without curriculum learning, using parameter scales from the noisy walking and quiet walking phases in \tabref{table:reward}. As shown in \figref{fig:tradeoff} (No Curriculum), only using the noisy walking phase parameters without curriculum achieves about half of the noise reduction compared to the proposed method. Using the quiet walking phase parameters in Table~\ref{table:reward} without curriculum, the training fails to converge. In other words, aibo doesn't learn to walk, due to too severe penalties for quiet walking. 

As a next ablation study, We tested the impact of removing switch contact sensors. Although the training converges, aibo doesn't walk at all. At the beginning of the quiet walking phase, aibo falls frequently, ending episodes early. We hypothesize that this is due to the lack of information about when to stiffen joints to support the whole body. Consequently, aibo learns to avoid walking to satisfy the strong noisy walking penalties.

As a third ablation study, a comparable result in terms of robustness is obtained using our proposed method without allowing the network to change the PD gains, as shown in \figref{fig:tradeoff} (Fixed PD gain). While this demonstrates that the noisy walking penalties and curriculum can reduce noise, the sound magnitude differs from our proposed method. PD gain scale change is beneficial for quiet walking, as mentioned in previous work \cite{bogdanovic2020learning}, because quiet walking is also a sensitive contact task.

Finally, we expanded our training of the proposed method by incorporating additional domain randomization parameters for ground characteristics, specifically the friction coefficient and terrain height. When modifying the friction coefficient range, increasing the maximum from 0.7 to 0.9 and decreasing the minimum from 0.4 to 0.2, we observed a relatively small sound magnitude but enhanced robustness compared to the originally proposed method as shown in \figref{fig:tradeoff} as More DR (terrain friction). Conversely, increasing the maximum terrain height from 0.01 to 0.03 resulted in the third-lowest sound magnitude while maintaining robustness comparable to the baseline as shown in \figref{fig:tradeoff} as More DR (terrain height). These findings suggest that the selection of domain randomization parameters enables tuning of the trade-off between robustness and quietness.

\begin{figure}
    \centering
    \vspace{2pt}
    \includegraphics[trim={0 0 0 0}, width=\linewidth]{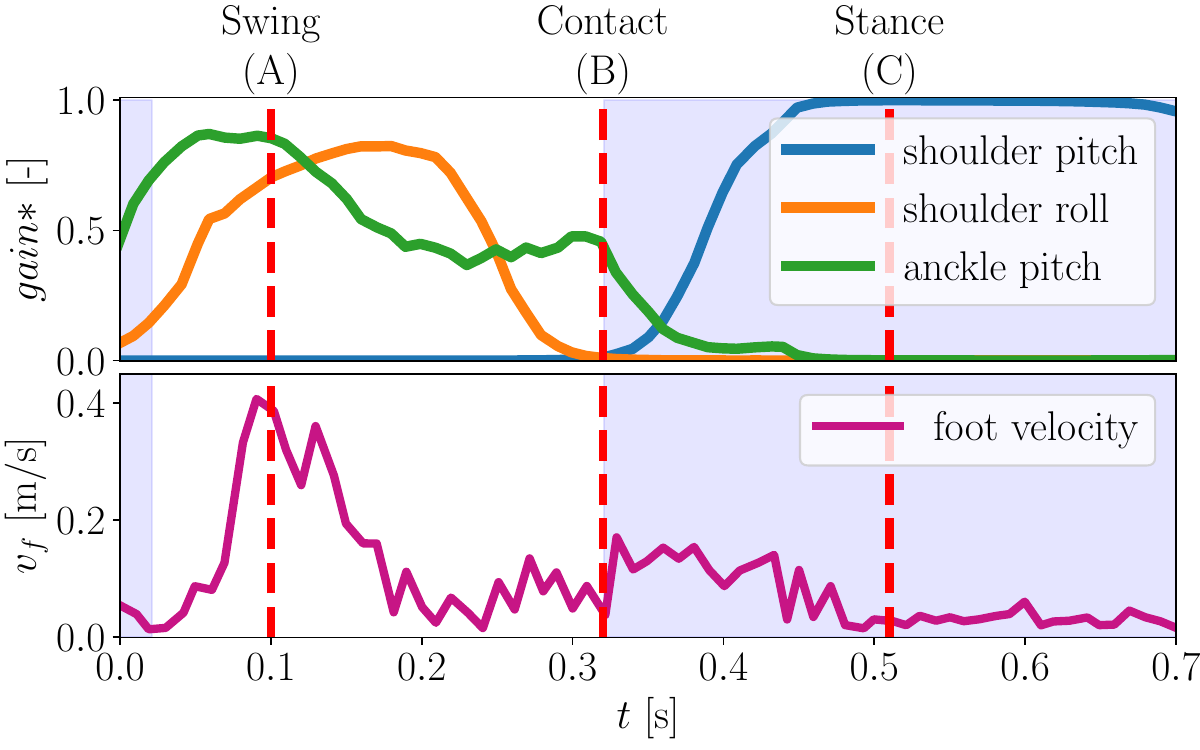}
    \caption{PD gain scale of $\sigma(x)$ in the \eqnref{eq:pd_gain_scale} at the fore right leg during locomotion. Right foreleg is (A): in the air to move forward, (B): approaching contact with the ground, (C): in contact to support the ensuing movement of the other legs.  The plot at the bottom shows the edge of the foot velocity. The blue area in the plot shows when the right foreleg is in contact with the ground, based on the foot switch contact sensor.}
    \label{fig:gain_scale_change}
    \vspace{-2ex}
\end{figure}

\subsection{PD Gain Scale Analysis}
\Figref{fig:gain_scale_change} illustrates the changes in PD gain scales and the edge of foot velocity exhibited by the quiet walking policy for the three actuators on the right foreleg. 
The plot of PD gain scale shows the scaled $\sigma(x_i)$ value outputted by the network, which is later scaled according to Equation~\ref{eq:pd_gain_scale}. 

As shown in the middle plot of \figref{fig:gain_scale_change}, just before the foot makes contact with the ground, the gains for all the actuators of the right foreleg decrease, effectively dampening the motion. This drop in gain scales at the shoulder pitch joint is followed by an increase, supporting the ensuing movement of the other legs. This behavior suggests that the policy learns to walk quietly by adjusting the PD gains to dampen the joint at the moment of ground contact.

The foot velocity, calculated through forward kinematics using measurement data of the three joint angles for the right foreleg, is shown in \figref{fig:gain_scale_change}. The foot velocity decreases just before contact with the ground. Conversely, when the foot is in the air, the velocity of the foot increases, along with the PD gains for shoulder roll and ankle pitch. This indicates that the policy actively controls the PD gains to maintain tracking of the base velocity.

%% file: sections/5-discussion.tex
\section{DISCUSSION}
While our proposed \ac{rl} policy demonstrates improved quietness, its robustness is not as high as the baseline \ac{rl} policy, resulting in reduced adaptability to unknown terrains. We have shown that the selection of \ac{dr} parameters, such as ground friction and terrain height, can influence the trade-off between robustness and quietness of the policy. Future work could incorporate the approach of Chen et al.~\cite{chen2024identifying}, which utilizes perception information to estimate ground characteristics like friction, height, and damping. This integration could enable the selection of an appropriate policy based on these factors, potentially allowing the robot to traverse slippery surfaces and employ robust walking strategies on various terrains.

Rather than directly minimizing footstep sound, which is challenging to emulate in physics simulator, we proposed sim-to-real \ac{rl} framework to minimize foot contact velocity, which correlates with footstep sound. This approach could be extended to other applications where real-world parameters are difficult to emulate in the typical locomotion simulator. For instance, battery life optimization could be addressed by minimizing torque in the simulator, as torque correlates with energy consumption. 

In this study, we focused primarily on reducing footstep sound during locomotion. However, other sources of noise during walking, such as actuator movement and mechanical friction, warrant further investigation. Although these sounds are not as dominant as footstep noise, addressing them remains crucial for developing quieter home-legged robots.

%% file: sections/6-conclusion.tex
\section{CONCLUSION}
In this work, we propose quiet walking based on a sim-to-real \ac{rl} approach to minimize foot contact velocity, which is highly correlated with footstep sound for a small home robot. Our framework consists of three key elements: adaptive PD gains, utilization of foot contact sensors, and implementation of curriculum learning. Through ablation studies, we demonstrate that each element contributes to minimizing the footstep sound for achieving a quieter walk. From the perspective of sound magnitude in the audible range from 20 Hz to 20 kHz, our approach achieves the quietest locomotion controller compared to Sony's commercial locomotion controllers and our own \ac{rl} baseline. We also identify a trade-off between quietness and robustness. 

Our work is the first to demonstrate and highlight key factors needed for the sim-to-real based \ac{rl} to achieve quiet walking. These findings have important implications for future research in human-robot interaction and the commercialization of quadruped robots for home use. 

% Our work is the first to demonstrate and highlight key factors necessary for an \ac{rl} based policy to achieve a quieter gait in quadruped robots. These findings have important implications for future research in human-robot interaction and the commercialization of quadruped robots for home use. By addressing the issue of noise generation during locomotion, our approach paves the way for more socially acceptable and less disruptive robotic companions in home environments.

%% file: acronyms.tex
\begin{acronym}
\acro{rl}[RL]{reinforcement learning}
\acro{ppo}[PPO]{Proximal Policy Optimization}
\acro{fft}[FFT]{Fast Fourier Transform}
\acro{dof}[DoF]{Degrees of Freedom}
\acro{mlp}[MLPs]{Multi-Layer Perceptrons}
\acro{dr}[DR]{Domain Randomization}
\acro{elu}[ELU]{Exponential Linear Unit}
\end{acronym}